\documentclass[journal,twoside,web]{ieeecolor}
\usepackage{tmi}
\usepackage{cite}
\usepackage{amsmath,amssymb,amsfonts, pifont}
\usepackage{algorithmic}
\usepackage{graphicx}
\usepackage{textcomp}
\usepackage{booktabs}
\usepackage{multirow}
\usepackage{tabularx}
\usepackage{threeparttable}

\newcommand{\cmark}{\ding{51}}%
\newcommand{\xmark}{\ding{55}}%

\def\BibTeX{{\rm B\kern-.05em{\sc i\kern-.025em b}\kern-.08em
    T\kern-.1667em\lower.7ex\hbox{E}\kern-.125emX}}
\begin{document}
\title{Enhancing medical vision-language contrastive learning via inter-matching relation modelling}
\author{Mingjian Li, Mingyuan Meng, Michael Fulham, David Dagan Feng, \IEEEmembership{Fellow, IEEE}, Lei Bi, \IEEEmembership{Member, IEEE}, and Jinman Kim, \IEEEmembership{Member, IEEE}
\thanks{Manuscript received Jan 19, 2024. This project was supported in kind by ARC DP200103748. \textit{(Corresponding authors: Jinman Kim.)}}
\thanks{Mingjian Li, Mingyuan Meng, Michael Fulham, David Dagan Feng, Lei Bi, and Jinman Kim are with the Biomedical data analysis and visualization lab, School of Computer Science, the University of Sydney, NSW, Australia (e-mail:mili3287@uni.sydney.edu.au; mmen2292@uni.sydney.edu.au; michael.fulham@sydney.edu.au; dagan.feng@sydney.edu.au; lei.bi@sydney.edu.au, jinman.kim@sydney.edu.au). }
\thanks{Michael Fulham is also with the Department of Molecular Imaging, Royal Prince Alfred Hospital, NSW, Australia (e-mail: michael.fulham@sydney.edu.au).}
\thanks{Lei Bi is also with the Institute of Translational Medicine, Shanghai Jiao Tong University, Shanghai, China (e-mail: lei.bi@sjtu.edu.cn).}}

\maketitle

\begin{abstract}
Medical image representations can be learned through medical vision-language contrastive learning (mVLCL) where medical imaging reports are used as weak supervision through image-text alignment. These learned image representations can be transferred to and benefit various downstream medical vision tasks such as disease classification and segmentation. Recent mVLCL methods attempt to align image sub-regions and the report keywords as local-matchings. However, these methods aggregate all local-matchings via simple pooling operations while ignoring the inherent relations between them. These methods therefore fail to reason between local-matchings that are semantically related, e.g., local-matchings that correspond to the disease word and the location word (semantic-relations), and also fail to differentiate such clinically important local-matchings from others that correspond to less meaningful words, e.g., conjunction words (importance-relations). Hence, we propose a mVLCL method that models the inter-matching relations between local-matchings via a relation-enhanced contrastive learning framework (RECLF). In RECLF, we introduce a semantic-relation reasoning module (SRM) and an importance-relation reasoning module (IRM) to enable more fine-grained report supervision for image representation learning. We evaluated our method using six public benchmark datasets on four downstream tasks, including segmentation, zero-shot classification, linear classification, and cross-modal retrieval. Our results demonstrated the superiority of our RECLF over the state-of-the-art mVLCL methods with consistent improvements across single-modal and cross-modal tasks. These results suggest that our RECLF, by modelling the inter-matching relations, can learn improved medical image representations with better generalization capabilities.
\end{abstract}


\section{Introduction}
\label{sec:introduction}

\begin{figure}[!t]
\centerline{\includegraphics[width=\columnwidth]{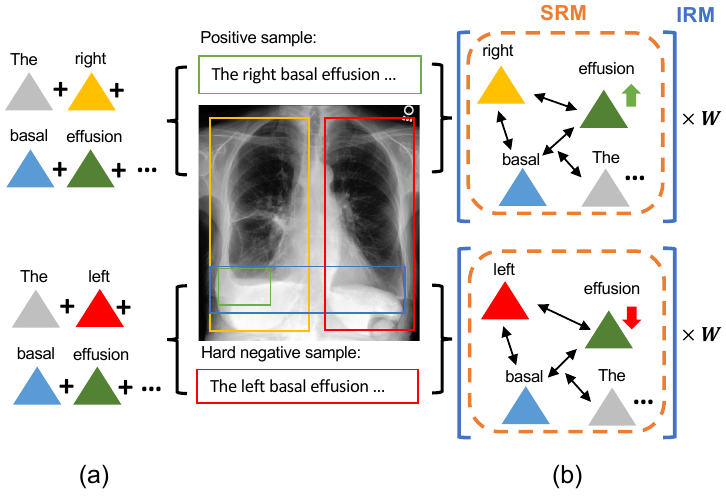}}
\caption{Illustration of commonly used mVLCL framework and our RECLF applied to a chest X-ray of a patient with a right basal effusion. One paired text sample (positive) and one unpaired text sample (hard negative) are shown as examples. (a) displays existing methods using a simple aggregation of local-matchings, and (b) displays our inter-matching relation-enhanced contrastive learning framework (RECLF). Triangles represent the local-matchings corresponding to the words on top and the image bounding box in same color. Black arrows show semantic-relations between local-matchings; green/red arrows show similarity variation of the local-matching after the semantic-relation reasoning module (SRM). The matrix $W$ represents the weight matrix of local-matchings of the importance-relation reasoning module (IRM).}
\label{fig1}
\end{figure}

\IEEEPARstart{M}{edical} vision-language contrastive learning (mVLCL) learns general medical image representations by leveraging the associated medical imaging  textual reports, as weak supervision, in a contrastive image-text alignment process. Zhang et al. \cite{zhang2022contrastive} reported the first mVLCL work, which was achieved by optimizing the latent medical image representations via a global-matching alignment between the paired medical images and the reports. This design provoked great interest in the community given the transferability of the learned image representations to the downstream tasks such as disease classification, lesion detection and segmentation \cite{tiu2022expertlevel,moor2023foundation,huang2023selfsupervised,krishnan2022selfsupervised,sun2023scoping}. However, critical clinical information sometimes can only occupy small sub-regions in medical images and the mVLCL methods that rely on global-matching are not able to capture these subtle findings \cite{NEURIPS2022_d925bda4,huang2021gloria}.

To achieve fine-grained image-text alignment, recent mVLCL methods tend to build and leverage the local-matchings between image sub-regions and report keywords \cite{huang2021gloria,NEURIPS2022_d925bda4}. Huang et al. \cite{huang2021gloria} proposed GLoRIA to automatically align important image sub-regions with each imaging report word and then sum all local-matchings to represent the entire finer-grained alignment. In addition, each local-matching can be viewed as a unit with strong relations that exist among the units (referred to as \textit{inter-matching relations}). On one hand, local-matchings are semantically related (\textit{semantic-relations}). As exemplified in Fig. \ref{fig1}, local-matchings corresponding to the words “right/left”, “basal”, and “effusion” are semantically related and can be aggregated to describe the location/presence of a basal effusion on a chest X-ray. On the other hand, each local-matching contains different grades of pathological information and thus is of unequal importance (\textit{importance-relations}), e.g., the word “The” in Fig. 1 does not contain useful pathological information and the corresponding local-matching is less important. These semantic- and importance-relations among local-matchings are crucial for mVLCL as they optimize the fusion of information from multiple semantic-related local-matchings and reinforce the important local-matchings. This enables more fine-grained and accurate image-text alignment. However, the inter-matching relations have not been explored by existing mVLCL methods, where local-matchings are treated in isolation and are aggregated together with simple pooling operations (Fig. 1a) \cite{huang2021gloria,liu2023improving}. Without considering the inter-matching relations, for example in Fig. 1, the positive sample “right basal effusion” and the hard negative sample “left basal effusion” are difficult to be differentiated. This is mainly because the local-matching corresponding to the disease word “effusion”, if not taken into context of its location word “right/left”, results in the highest similarity measurement in both cases. This inevitably results in less differentiable image-text alignment and thus hinders representation learning.

In this study, we attempt to enhance mVLCL by explicitly modelling the inter-matching relations among local-matchings. Our contributions are as follows:
\begin{itemize}
    \item a relation-enhanced contrastive learning framework (RECLF, Fig. 1b) to model both the semantic- and the importance- relations, thus enabling more fine-grained report supervision for image representation learning;
\end{itemize}
\begin{itemize}
    \item a semantic-relation reasoning module (SRM) to model high-level semantic-relations among local-matchings, thus allowing more accurate similarity measurements;
\end{itemize}
\begin{itemize}
    \item an importance-relation reasoning module (IRM) to model the importance-relations among local-matchings, thus enabling the model to focus on critical local-matchings guided by the text encoder.
\end{itemize}

We conducted extensive experiments with six public datasets on four downstream tasks including segmentation, zero-shot classification, linear classification, and cross-modal retrieval.

\section{Related Work}
\subsection{Medical Vision-Language Contrastive Learning (mVLCL)}
Medical vision-language contrastive learning (mVLCL) \cite{zhang2022contrastive} is an emerging technique that allows to learn image representations that can be transferred to and benefit various downstream tasks. The image representations are usually learned through a “pseudo instance-level classification task” where the text is employed as the “pseudo label”. More specifically, a latent representation space can be determined, when the representation similarities between the image and its paired text report (“pseudo label”) are maximized while the representation similarities between the image and unpaired text reports are minimized. By this means, the semantics from the text can implicitly guide the image representations to be differentiated for different classes in the latent space.

Zhang et al. \cite{zhang2022contrastive} first aligned the images and text reports at the global level where one single global representation was extracted for each image and imaging text report, respectively. The global-matching between the image and the text report is stringent, considering a positive match only when they come from the same patient (paired image-text). However, the unpaired medical images and imaging reports could also have large similarities, e.g., they depict the same disease. Other methods \cite{NEURIPS2022_d925bda4,liu2023improving,wang2022medclip,chen2023knowledge} proposed to exploit a more soft matching by considering all images and text reports containing similar semantics as positive matches.

However, these global-matching methods are limited when the important semantic pathological information only occupies a small proportion of the image and the surrounding background regions are irrelevant. To address this limitation, Karpathy et al. \cite{karpathy2015deep} proposed to use a pretrained detection model to extract ROIs in natural images and then match the ROIs with words in the text. Due to the lack of pretrained object detection model for medical imaging related research, Huang et al. \cite{huang2021gloria} and Müller et al. \cite{muller2022joint} proposed to treat the representation of each pixel in the last convolution layer of ResNet \cite{he2016deep} as the localized image representations of the image sub-regions and matched with the representations of the word or the sentence at the local level. Recently, the visual transformer (ViT) \cite{dosovitskiy2020image} has shown great success in various medical image analysis tasks \cite{shamshad2023transformers}, which naturally splits the image into fixed-size non-overlapping image patches. Motivated by this great success, Wang et al. \cite{NEURIPS2022_d925bda4} and Chen et al. \cite{chen2022multimodal} employed ViT as the image encoder and intuitively identified the most relevant image patches corresponding to the word in text reports via a cross-attention module. Such fine-grained image-text local-matching has achieved performance improvements. However, these methods attempt to view the image and text as separate entities, such that only the image-text relation within each local-matching is explored. Consequently, they do not explore the relations among different local-matchings (\textit{inter-matching relations}), potentially making it difficult to distinguish the challenging negative samples.

\subsection{Relation Modelling}
Medical images contain a vast amount of information, e.g., multiple structures, lesions, etc., and the relation modelling helps clinicians to understand the global context. The idea to model the relations and to promote the information passing in deep learning models has seen improvements in various medical image analysis tasks, such as in disease prediction \cite{ghorbani2022ragcn}, lesion detection \cite{li2022satr}, structure segmentation \cite{li2022dual}, and image registration \cite{nakao2022imagetograph}.

Relation modelling has also been explored in vision-language tasks. For feature extraction, Transformer \cite{vaswani2017attention} has shown great capability to model long-range relations for text and images. BERT \cite{devlin2019bert} and its variants, e.g., BioClinicalBERT \cite{alsentzer2019publicly} and CXR-BERT \cite{boecking2022making} which are pretrained on medical text, have been widely employed by mVLCL methods as domain-specific language encoders. Moreover, Wu et al. \cite{wu2023medklip} and Zhang et al. \cite{zhang2023knowledgeenhanced} proposed to extract medical terminology triplets from the raw text input using an Named Entity Recognition method RadGraph \cite{jain2021radgraph}. The intra-modality text-text semantic relations were explicitly encoded and the unnecessary complexity caused by the writing styles was removed. Vision Transformer (ViT) \cite{dosovitskiy2020image} was also adopted to extract image representations in mVLCL \cite{NEURIPS2022_d925bda4,zhao2023clip,wan2023medunic}, which enables the exploration of relations spanning the entire image. Wang et al. \cite{NEURIPS2022_d925bda4} further employed an additional self-attention module to model the patch-to-patch relations within the images. These methods explored relations within each modality, e.g., text or image. Additionally, relations modelling has been conducted subsequent to feature extraction. In the natural image-text retrieval task, Liu et al. \cite{liu2020graph} and Diao et al. \cite{diao2021similarity} investigated the potential to merge information from local-matchings into global-matching. Huang et al. \cite{huang2021gloria} employed a cross-attention module to identify the important image patches based on their significance for a given word, and vice versa. Chen et al. \cite{chen2022multimodal} proposed a multimodal self-attention module to fuse the information from images and text. These methods explored the relations across tokens of different modalities. Unfortunately, the potentials of viewing of each image-text local-matching as an integrated unit and the relations among them have not been explored in mVLCL. We propose to extend mVLCL by incorporating inter-matching relation modelling to obtain more fine-grained guidance from the text for image representation learning. 

\begin{figure*}[!t]
\centerline{\includegraphics[width=\textwidth]{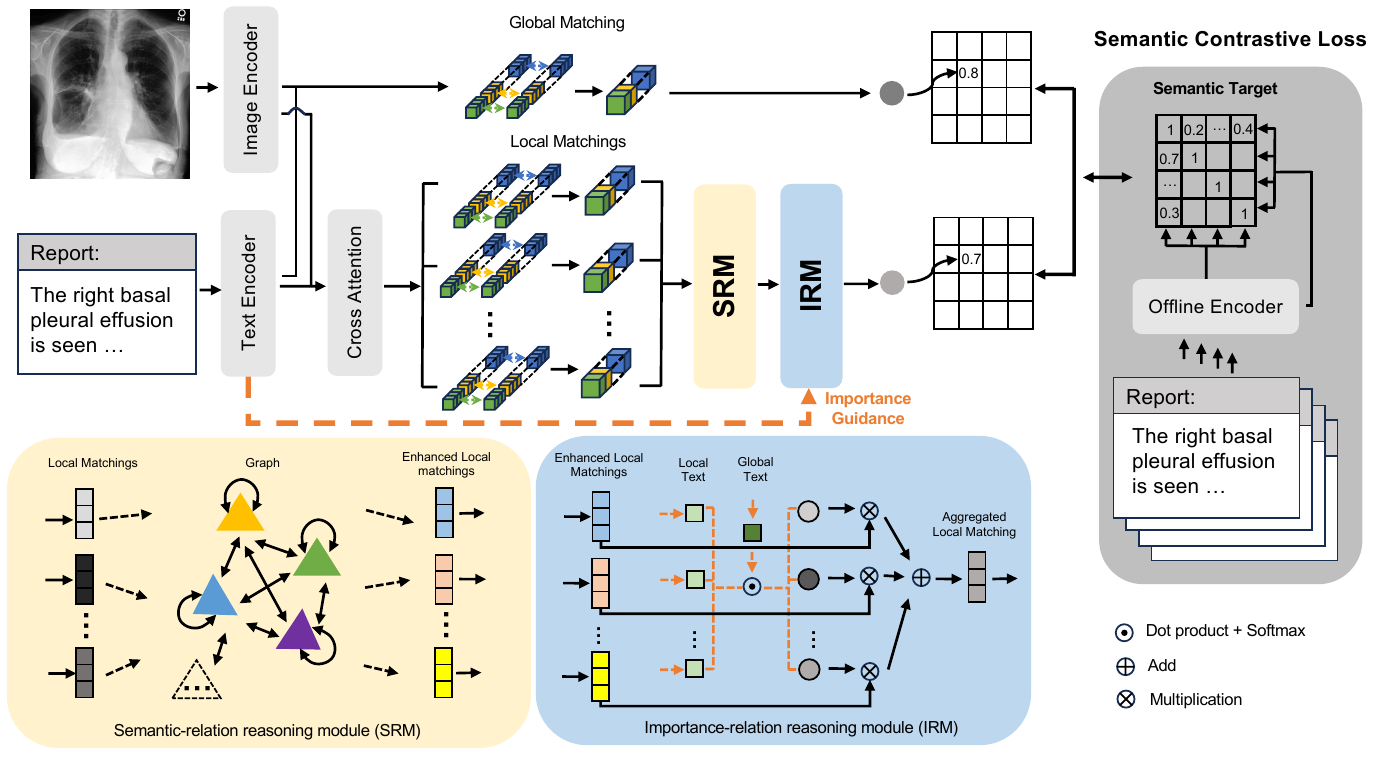}}
\caption{An overview of our RECLF. RECLF extracts representations of images and text and then calculates the similarities of the global- and local-matchings. SRM propagates the semantic information among all local-matchings and IRM aggregates all local-matchings with the importance guidance from the text encoder. Semantic contrastive loss is then applied to both the global and the local branches.}
\label{fig2}
\end{figure*}

\section{Methodology}
Our RECLF is illustrated in Fig. 2,   which consists of three representation extraction modules (image encoder, text encoder, and cross-attention module, detailed in Section III-A) and two relation reasoning modules (SRM and IRM, detailed in Sections III-B and III-C). The framework is trained with a semantic contrastive loss as outlined in Section III-D.

\subsection{Representation Extraction}
We used the commonly adopted image encoder - ViT-B/16 to extract image representations. The input images were split into $16\times16$ non-overlap image patches, which were then flattened into \textit{M} patch tokens. We followed the standard implementation to add position embeddings to the patch tokens, and the derived patch tokens were sent through the transformer blocks. We extracted the global image representation $I_g \in \mathbb{R}^{768}$ from the [CLS] token of the image encoder. We also extracted the local image representation $I_l \in \mathbb{R}^{768 \times M}$ from patch tokens, which represents \textit{M} sub-regions on the original input image.

We used the BioClinicalBERT as the text encoder to extract text representations. Specifically, we extracted \textit{N} word representations from the encoder as $T_l \in \mathbb{R}^{768 \times N}$. The global text representation is defined as the sum-up of all the \textit{N} word representations $T_{g}=\sum_{i=1}^N{T_{l,i}},T_g \in \mathbb{R}^{768}$. 

The global-matching represents the alignment between the global image and text representations, $T_{g}$ and $I_{g}$. For local-matchings, we employed a cross-attention module to correlate the most related image sub-regions with each textual word. We first computed the dot-product similarity between all local image representations and a specific word representation:
\begin{equation}
    s_{i}=I_{l}^T T_{l,i}   
\end{equation}
where $s_i \in \mathbb{R}^M$ indicates the similarity between all the $M$ local image representations and the $i \in[0, N]$ word representation. We then calculated the cross-attention weight $a_{i,j}$ of $j \in[0, M]$ local image representation based on its normalized similarity with \textit{i}-th word, which is denoted as follows:
\begin{equation}
    a_{i,j} = \frac{exp(s_{i,j}/\tau_{1})}{\sum_{m=1}^M exp(s_{i,m}/\tau_{1})}
\end{equation}
where $\tau_{1}$ is a temperature hyper-parameter. The cross-attention-weighted local image representation $V_{l, i} \in \mathbb{R}^{768}$ for the \textit{i}-th word can be generated as:
\begin{equation}
    V_{l,i} = \sum_{j=1}^M a_{i,j}I_{l,j}
\end{equation}
The local-matchings represent the alignment between each word representation $T_{l, i} \in \mathbb{R}^{768}$ and its cross-attention-weighted local image representations $V_{l,i}$.

\subsection{Semantic-relation Reasoning Module}
Our SRM first calculates the similarity of each local-matching, then explicitly models the semantic-relations among all local-matchings, and finally propagates and aggregates the information among all local-matchings. For the similarity calculation, existing mVLCL methods usually calculated the scalar cosine distance or Euclidean distance between the image sub-region and text word of each local-matching. In contrast, we calculated the similarity in a more fine-grained manner. We first split the \textit{i}-th word representation $T_{l,i}$ and its cross-attention-weighted local image representation $V_{l,i}$ into \textit{k} blocks along the channel dimension, denoted by $[t_{l,i1},t_{l,i2},...,t_{l,ik};t_{l,is}\in\mathbb{R}^{\frac{768}{k}},s\in[0,k]]$ and $[v_{l,i1},v_{l,i2},...,v_{l,ik};v_{l,is}\in\mathbb{R}^{\frac{768}{k}},s\in[0,k]]$. We then calculated the scalar cosine similarity $\cos(\ast)\in\mathbb{R}$ between each paired block and concatenated them as the vectorized similarity $s_i'$ of \textit{i}-th local-matching: 
\begin{equation}
    s_i'=[cos(t_{l,i1},v_{l,i1}),cos(t_{l,i2},v_{l,i2}),\ldots, cos(t_{l,ik},v_{l,ik})]
\end{equation}
where $s_{i}^{\prime}\in\mathbb{R}^{k}$. We also calculated the vectorized similarity between the global representations of image and text of global-matching as $s_{g}^{\prime}\in\mathbb{R}^{k}$.

For semantic-relation modelling, we constructed a complete directed graph with \textit{N} nodes where each node represents a local-matching and is connected to all other \textit{N}-1 nodes. We follow the graph attention network \cite{velickovic2018graph} to assign edge weights, through a cross-attention mechanism, to represent the semantic relationships among different local-matchings. To ensure sufficient discriminative power for transforming input features (similarity vectors) into higher-level features, the nodes were initially projected using shared fully-connected layers $f$. We then applied self-attention mechanisms to the nodes, where the derived attention coefficients can indicate the significance of node $x$ to node $y$. Specifically, the directed edge $E_{x\rightarrow y}$ from the node $x$ to the node $y$ is defined as:

\begin{equation}
   E_{x\rightarrow y} = \frac{exp((f_{x}s_{x}')(f_{y}s_{y}'))}{\sum _x exp((f_{x}s_{x}')(f_{y}s_{y}'))}
\end{equation}
where $f_{x}(k\rightarrow k)$ and $f_{y}(k\rightarrow k)$ are two trainable linear transformation functions. For the semantic information propagation and aggregation, we computed a linear combination of corresponding features by multiplying them with normalized coefficients $E$, which serves as the final output features for each node. This results in a semantic-enhanced vectorized similarity $s_{i}^{''}\in\mathbb{R}^{k}$ that allows to capture information from the current local-matching and all other semantic-related local-matchings.

\subsection{Importance-relation Reasoning Module}
Our IRM models the importance-relations among local-matchings and then selectively attends to the most important local-matchings after SRM. For the importance-relations modelling, we employed the guidance from the text encoder. Specifically, we calculated the normalized dot product similarity between the \textit{i}-th word representation $T_{l,i}\in\mathbb{R}^{768}$ and the global text representation $T_g$, which can be defined as:
\begin{equation}
    \omega _i = T_{l,i}^T T_g
\end{equation}
Then, we normalized $\omega_{i}\in\mathbb{R}$ across the text word dimension using the Softmax with temperature $\tau_2$. For the selection and prioritization of the important local-matchings, we aggregated all the semantic-enhanced vectorized similarity $s_i''$ with the importance guidance value $\omega _i$ and then mapped them into a scalar similarity value $\hat{s}_{l_{irm}}\in\mathbb{R}$ using a fully connected layer $g(k\rightarrow1)$:
\begin{equation}
    \hat{s}_{l_{irm}} = g(\sum_i \omega _i s_i'')
\end{equation}
We also mapped the global similarity $s_g'$ to a single value ${\hat{s}}_g$ using the same fully connected layer \textit{g}. The ${\hat{s}}_g$ and ${\hat{s}}_{l_{irm}}$ are the final similarity measurements between a pair of image and text at the global and local level, respectively.

\subsection{Semantic Contrastive Loss}
The learning objective of our RECLF is to maximize the similarity between the semantically matched medical image-text pairs while minimizing the similarity between the non-matched pairs. Since the text reports provide richer semantics with more notable differences than images \cite{liu2023improving}, we took the similarity between the text reports as the soft semantic target, which was defined in \cite{wang2022medclip}. More specifically, we employed CheXpert to build the multi-label class label from text associated with the \textit{p}-th image-text pair, denoted as $l_{p}\in\mathbb{R}^{z}$ where $z$ is the number of classes. Therefore, the soft semantic target $\bar{s}\in\mathbb{R}$ for any sampled image (from \textit{p}-th pair) and text (from \textit{q}-th pair) during the training stage is defined by the cosine similarity between the multi-hot label vector $l_p$ and $l_q$.

Following MedCLIP \cite{wang2022medclip}, we employed the semantic contrastive loss as the objective function to reduce false negatives, which is calculated as a cross-entropy loss between the predicted global similarity, SRM and IRM enhanced local similarity, and the soft semantic target, respectively:
\begin{equation}
    L=-\frac{1}{B z}\sum_{p=1}^{B z}\sum_{q=1}^{B z}\left(\bar{s}_{p,q}*\log\left(\hat{s}_{g_{p,g}}\right)+\bar{s}_{p,q}*\log\left(\hat{s}_{l_{irm_{p,q}}}\right)\right)
\end{equation}
where \textit{Bz} is the batch size. Note that we computed the cross-entropy loss in both image-to-text direction and text-to-image direction and averaged them to derive the final loss.

\section{Experimental Setup}
\subsection{Pre-training} 
\subsubsection{Dataset} We pre-trained our method with MIMIC-CXR \cite{johnson2019mimiccxr} which is a public dataset curated by the Beth Israel Deaconess Medical Center, Boston, MA, USA. This dataset contains over 220K   AP-view (AP refers to Antero-Posterior) chest X-rays and their associated text reports. Following GLoRIA \cite{huang2021gloria}, We randomly sampled a MIMIC-5x200 dataset (a total of 1000 image-text pairs) for downstream tasks including cross-modal retrieval and zero-shot classification. Another 5000 random samples were held out for validation and the rest were used for training. 

\subsubsection{Implementation details} All images were scaled to 256 × 256, and image augmentation was applied by randomly cropping to 224 × 224. For the text input, we used the “findings” and “impression” sections of the report, consistent with MGCA \cite{NEURIPS2022_d925bda4}. We used ViT-B/16 as our image encoder backbone and pre-trained the BioClinicalBERT \cite{alsentzer2019publicly} as our text encoder. We summed up the outputs from the last 4 layers of the text encoder as the representation of each text word. All representations were projected into a common latent space with a dimension of 768. For pre-training, we employed an AdamW optimizer with a learning rate of $1e^{-5}$ and a weight decay of 0.01. We trained our method for 100K steps using a Nvidia 24GB RTX3090 GPU with a batch size of 48. We used cosine annealing scheduler with linear warmup. The initial learning rate was set to 0 and the warmup steps were 10K. We set the temperature hyperparameters to $\tau_1=4,\tau_2=5$.

\subsection{Downstream Task 1: Segmentation} 
\subsubsection{Dataset} We evaluated the segmentation task using the SIIM Pneumothorax \cite{siim-acr-pneumothorax-segmentation}, RSNA Pneumonia dataset \cite{shih2019augmenting}, and TBX11K dataset \cite{liu2020rethinking}. The SIIM dataset comprises 12K chest X-rays, each with a manually annotated mask delineating a pneumothorax. The RSNA dataset contains over 29K chest X-rays and associated bounding boxes indicating evidence of pneumonia. Following MGCA \cite{NEURIPS2022_d925bda4}, we split the SIIM and RSNA datasets into train/validation/test sets with a distribution of 70\%/15\%/15\%. The TBX11K dataset contains 11K X-ray images with corresponding bounding boxes for tuberculosis. We directly used its official validation set as our test set and sampled 15\% data of its official train set for validation, using the rest for training.

\subsubsection{Implementation Details} We employed the well-established segmentation transformer (SETR) \cite{zheng2021rethinking} to generate segmentation predictions. We replaced the SETR’s image encoder with our pretrained image encoder and fixed it. The rest of the SETR (decoder) was trained with either 1\%, 10\%, and 100\% of the training data. We employed the AdamW optimizer with a learning rate of $5e^{-4}$ and a weight decay of 0.05. We used a cosine annealing scheduler. The segmentation performance was evaluated with the Dice similarity coefficient. 

\subsection{Downstream Task 2-3: Linear and Zero-shot Classification}
\subsubsection{Dataset} Two classification tasks were evaluated: \textit{linear} and \textit{zero-shot}. We first evaluated linear classification (supervised) on 4 well-benchmarked datasets: (1) CheXpert dataset \cite{irvin2019chexpert}, which comprises over 190K chest X-rays associated with five binary labels: atelectasis, cardiomegaly, consolidation, edema, and pleural effusion. Following MGCA \cite{NEURIPS2022_d925bda4}, the raw validation set was used for testing. A randomly selected 5,000 samples from the raw training set were held out for validation and the rest were used for training. (2) RSNA Pneumonia dataset \cite{shih2019augmenting}, which contains binary labels of normal and pneumonia. (3) NIH dataset \cite{wang2017chestxray8}, which compromises 112K images with 14 disease labels. We sampled 10\% data from its training set for validation. (4) SIIM Pneumothorax dataset \cite{siim-acr-pneumothorax-segmentation}, which contains binary labels of normal and pneumothorax.

We also evaluated our learned model with zero-shot classification on the MIMIC-5x200 dataset which consists of 200 exclusively positive images for each disease category (5 categories: atelectasis, cardiomegaly, consolidation, edema, pleural effusion). The text reports were replaced with 5 × 5 caption prompts (5 captions for each class) from GLoRIA \cite{huang2021gloria} created by a radiologist.

\subsubsection{Implementation Details} For the linear classification task, we froze the pre-trained image encoder and only trained a one-layer linear classification head following MGCA \cite{NEURIPS2022_d925bda4}. We evaluated our method with 1\%, 10\%, and 100\% training data. We employed the AdamW optimizer with a learning rate of $5e^{-4}$ and a weight decay of $1e^{-6}$. We used area under the ROC curve (AUROC) as the evaluation metric.

For the zero-shot classification task, we took an image as the input and predicted its class label by matching it with the text prompts of each class. Specifically, we encoded the input image and text prompts from 5 classes using our pre-trained image and text encoders. The input image was predicted as the class whose text prompts have the highest average similarity with the image. We used AUROC, Accuracy, Precision, and F1 score as the evaluation metric.

\subsection{Downstream Task 4: Cross-modal Retrieval} 
\subsubsection{Dataset} We employed the same MIMIC-5x200 dataset from MIMIC-CXR for cross-modal retrieval.

\subsubsection{Implementation Details} 
For image-to-text retrieval, we used an image as the query and retrieved the relevant text; similarly, for text-to-image, we used text as the query and retrieved relevant images. As the evaluation metrics, we employed Precision@K (P@K) by calculating the precision in the top K={1,5,10} retrieved results. The higher P@K indicates better performance. To show the overall retrieval performance, we computed the sum of all P@K of two directions as P@Sum.

\subsection{Comparison Methods}
For single-modal comparisons that only involved the learned image encoders, including segmentation and linear classification, we compared with the state-of-the-art mVLCL methods including ConVIRT \cite{zhang2022contrastive}, GLoRIA \cite{huang2021gloria}, MGCA \cite{NEURIPS2022_d925bda4}, MedCLIP \cite{wang2022medclip}, SAT \cite{liu2023improving} and MedKLIP \cite{wu2023medklip}. ConVIRT \cite{zhang2022contrastive} focused on learning the global-matching between the entire image and text, while GLoRIA \cite{huang2021gloria} identified the most important image sub-regions for a given word and exploited the local-matchings between them. MGCA \cite{NEURIPS2022_d925bda4}, MedCLIP \cite{wang2022medclip}, and SAT \cite{liu2023improving} leveraged high-level semantic (disease) correspondences. Specifically, MGCA \cite{NEURIPS2022_d925bda4} enforced the consistency between the self-clustering results of the image and text representations. MedCLIP \cite{wang2022medclip} used a soft semantic loss to incorporate the disease-level information into the image-text global-matching. SAT \cite{liu2023improving} calculated the contrastive loss on fine-grained image-text matchings which were classified into positive, negative, and neutral according to their disease correspondences. MedKLIP \cite{wu2023medklip} simplified the text reports into a set of semantic entity triplets as \{entity, position, exist\} and exploited the text guidance at the entity level. Recently, generative-based medical vision-language pretraining has also shown great capability in learning image representations by reconstructing the masked patches using the context \cite{he2021masked}. One of such generative-based methods named MRM \cite{zhou2023advancing}, and another work named M3AE \cite{chen2024mapping} which combines mVLCL with generative method were also included for comparison.

We grouped the existing mVLCL methods into two categories according to the used image encoder: \textit{CNN-based} or \textit{Transformer-based}. We also included a widely used \textit{image-only} self-supervised learning method – MoCo v2 \cite{chen2020improved} to demonstrate the effectiveness of incorporating text as guidance in image representation learning. Finally, we included the baseline image encoders that were either randomly initialized or pretrained with ImageNet, to demonstrate the effectiveness of learning image representations tailored for medical images.

For all comparison methods, we used their officially released pretrained model weights and evaluated them using the same experimental settings from MGCA to ensure a fair comparison. This setting caused changes to the fine-tuning of some of the comparison methods, such as MRM \cite{zhou2023advancing} (excluded fine-tuning of the encoder).

\section{Results}
\subsection{Segmentation}
The segmentation results are shown in Table I. Overall, methods using Transformer as the image encoder outperformed CNN-based methods. Our method consistently outperformed other methods in Dice under different data ratios across three datasets. In addition, our method with only using 1\% of fine-tuning data outperformed the state-of-the-art MRM by a large margin of 4.6\% in Dice on the SIIM dataset. 

\begin{table*}[t]
\centering
\caption{Segmentation Results on the SIIM, RSNA, and TBX11K Datasets}
\label{table1}
\setlength{\tabcolsep}{16pt}
\begin{threeparttable}
\begin{tabular}{lccc|ccc|cc}
\toprule
\multirow{2}{*}{Methods}            & \multicolumn{3}{c|}{SIIM}                      & \multicolumn{3}{c|}{RSNA}                      & \multicolumn{2}{c}{TBX11K}    \\
                                    & 1\%           & 10\%          & 100\%         & 1\%           & 10\%          & 100\%         & 10\%          & 100\%         \\
\midrule
Random Init                         & 9.0             & 28.6          & 54.3          & 6.9           & 10.6          & 18.5          & 15.7          & 25.3          \\
ImageNet   Init                     & 10.2          & 35.5          & 63.5          & 34.8          & 39.9          & 64.0            & 37.8          & 58.0            \\
\midrule
\textit{\textbf{CNN-based}}         &               &               &               &               &               &               &               &               \\
ConVIRT \cite{zhang2022contrastive}                             & 25.0            & 43.2          & 59.9          & 55.0            & 67.4          & 67.5          & 52.4          & 60.9          \\
GLoRIA-MIMIC \cite{huang2021gloria}                      & 37.4          & 57.1          & 64.0           & 60.3          & 68.7          & 68.3          & 55.1          & 62.6          \\
MedCLIP \cite{wang2022medclip}                             & 51.2          & 62.3          & 67.8          & 66.5          & 69.6          & 71.3          & 58.2          & 66.9          \\
MGCA-ResNet \cite{NEURIPS2022_d925bda4}                        & 49.7          & 59.3          & 64.2          & 63.0            & 68.3          & 69.8          & 55.9          & 64.2          \\
MedKLIP* \cite{wu2023medklip}                        & 42.5          & 61.9          & 66.8          & 64.7            & \underline{71.2}          & \textbf{73.3}          & 60.6          & 65.8          \\
SAT \cite{liu2023improving}                                  & 47.6          & 57.8          & 64.2          & 68.5          & 70.6          & 71.1          & 56.9          & 65.0            \\
\midrule
\textit{\textbf{Transformer-based}} &               &               &               &               &               &               &               &               \\
GLoRIA-ViT \cite{huang2021gloria}                        & 48.2          & 56.9          & 64.7          & 62.6          & 68.4          & 69.2          & 58.6          & 64.7          \\
PTUnifier   \cite{Chen_2023_ICCV}                & 49.1          & 57.1          & 64.0            & 64.8          & 68.6          & 70.5          & 60.9          & 66.3          \\
REFERS   \cite{zhou2022generalized}                   & 50.6          & 58.9          & 64.5          & 64.6          & 68.1          & 70.3          & 61.4          & 66.7          \\
MGCA-ViT \cite{NEURIPS2022_d925bda4}                          & 51.3          & 57.3          & 65.4          & 68.8          & 69.2          & 70.9          & 62.5          & 66.8          \\
MRM \cite{zhou2023advancing}                                 & \underline{62.3}    & \underline{67.5}    & 68.6          & \underline{69.3}    & 70.7          & \underline{71.7}    & \underline{63.2}    & \underline{68.2}    \\
M3AE   \cite{chen2024mapping}                 & 62.0            & 67.4          & \underline{68.8}    & 69.0            & 70.1          & 70.7          & 62.8          & 67.3          \\
Ours                                & \textbf{66.9} & \textbf{68.5} & \textbf{69.3} & \textbf{72.1} & \textbf{72.3} & \textbf{73.3} & \textbf{63.3} & \textbf{68.5}\\
\bottomrule
\end{tabular}
\begin{tablenotes}
    \footnotesize
    \item {The segmentation results are measured using Dice similarity coefficient (\%). The \textbf{best} results are bolded and the \underline{second-best} results are underlined. *: MedKLIP pretrained weights do not contain the last residual block and was randomly initialized.}
\end{tablenotes}
\end{threeparttable}
\end{table*}

\subsection{Zero-shot and Linear Classification}
Table II shows the zero-shot classification results. Our method consistently outperformed all other methods across all 5 metrics with a large margin of ~4\% on AUROC, ~7\% on Accuracy, ~5\% on Precision, and ~7\% on F1 score.

The results of linear classification are shown in Table III. Our method achieved a better performance in 6 out of 12 metrics. Our method also outperformed the image-only self-supervised learning method MoCo V2 by a large margin of 6.4\% with only using 1\% of CheXpert data and by 4.3\% with only using 1\% of RSNA data.

\begin{table}[]
\centering
\caption{Zero-shot Classification Results on the MIMIC-5x200 Dataset}
\label{table2}
\setlength{\tabcolsep}{7pt}
\begin{threeparttable}
\begin{tabular}{lcccc}
\toprule
\multirow{2}{*}{Methods} & \multicolumn{4}{c}{MIMIC-5x200}     \\
                         & AUROC & Accuracy & Precision & F1   \\
                         \midrule
Random Init              & 0.47  & 0.20     & 0.04      & 0.07 \\
ImageNet Init            & 0.51  & 0.20     & 0.06      & 0.07 \\
\midrule
\textbf{\textit{CNN-based}}                &       &          &           &      \\
ConVIRT \cite{zhang2022contrastive}          & 0.81  & \underline{0.60}     & 0.60      & \underline{0.60} \\
GLoRIA-MIMIC \cite{huang2021gloria}     & 0.81  & 0.54     & 0.57      & 0.55 \\
SAT \cite{liu2023improving}             & 0.79  & 0.53     & 0.53      & 0.52 \\
\midrule
\textbf{\textit{Transformer-based}}       &       &          &           &      \\
GLoRIA-ViT \cite{huang2021gloria}       & 0.80  & 0.53     & 0.54      & 0.53 \\
MGCA-ViT \cite{NEURIPS2022_d925bda4}         & \underline{0.84}  & \underline{0.60}     & \underline{0.62}      & 0.59 \\
Ours                     & \textbf{0.88}  & \textbf{0.67}     & \textbf{0.67}      & \textbf{0.67} \\
\bottomrule
\end{tabular}
\begin{tablenotes}
    \footnotesize
    \item {The \textbf{best} results are bolded and the \underline{second-best} results are underlined.}
\end{tablenotes}
\end{threeparttable}
\end{table}

\begin{table*}[t]
\centering
\caption{Linear Classification Results on the CheXpert, RSNA, NIH, and SIIM Datasets}
\label{table3}
\setlength{\tabcolsep}{9pt}
\begin{threeparttable}
\begin{tabular}{lccc|ccc|ccc|ccc}
\toprule
\multirow{2}{*}{Methods} & \multicolumn{3}{c|}{CheXpert}                  & \multicolumn{3}{c|}{RSNA}                      & \multicolumn{3}{c|}{NIH}                       & \multicolumn{3}{c}{SIIM}                      \\
                         & 1\%           & 10\%          & 100\%         & 1\%           & 10\%          & 100\%         & 1\%           & 10\%          & 100\%         & 1\%           & 10\%          & 100\%         \\
\midrule
Random Init                  & 56.1          & 62.6          & 65.7          & 58.9          & 69.4          & 74.1          & 50.0            & 53.6          & 57.7          & 44.5          & 55.4          & 58.3          \\
ImageNet Init                 & 74.4          & 79.7          & 81.4          & 74.9          & 74.5          & 76.3          & 58.1          & 69.1          & 79.0            & 56.9          & 73.2          & 79.4          \\
\midrule
\multicolumn{13}{l}{\textit{\textbf{Image-only}}}                                                                                                                                                                        \\
MoCo V2 \cite{chen2020improved}                  & 81.6          & 84.7          & 84.9          & 85.0            & 87.4          & 88.2          & -             & -             & -             & -             & -             & -             \\
\midrule
\multicolumn{13}{l}{\textit{\textbf{CNN-based}}}                                                                                                                                                                         \\
ConVIRT \cite{zhang2022contrastive}                 & 85.9          & 86.8          & 87.3          & 77.4          & 80.1          & 81.3          & 64.9          & 77.1          & 80.8          & 84            & 85.6          & 87.6          \\
GLoRIA-MIMIC  \cite{huang2021gloria}                  & 87.1          & 88.7          & 88.0            & 87.0            & 89.4          & 90.2          & 59.7          & 74.3          & 80.0           & 86.0            & 86.7          & 88.5          \\
MedCLIP   \cite{wang2022medclip}                & 87.1          & 87.4          & 88.1          & 88.3          & 89.3          & 90.1          & 60.9          & 74.8          & 80.1          & 87.3          & 88.0            & 89.3          \\
MGCA-ResNet  \cite{NEURIPS2022_d925bda4}           & 87.6          & 88.0            & 88.2          & 88.6          & 89.1          & 89.9          & 72.0            & 79.3          & 82.4          & 85.6          & 86.5          & 88.8          \\
MedKLIP* \cite{wu2023medklip}                   & 77.7          & 82.7          & 84.5          & 83.6          & 86.8          & 87.8          & 60.9          & 74.8          & 80.1          & 86.0            & 88.6          & 90.8          \\
SAT \cite{liu2023improving}                       & 86.9          & 88.3          & 88.6          & 87.4          & 89.2          & 90.2          & 68.5          & 75.1          & 80.4          & 85.3          & 87.0            & 89.0            \\
\midrule
\multicolumn{13}{l}{\textit{\textbf{Transformer-based}}}                                                                                                                                                                 \\
GLoRIA-ViT \cite{huang2021gloria}            & 86.2          & 88.4          & 88.2          & 87.4          & 89.4          & 90.6          & 71.5          & 78.4          & 81.6          & 87.8          & 89.3          & 90.2          \\
PTUnifier \cite{Chen_2023_ICCV}       & \underline{88.7}    & 89.0            & \textbf{90.1} & 88.7          & 89.5          & 90.6          & 75.1          & 80.2          & 81.9          & 87.9          & 91.6          & 92.0            \\
REFERS \cite{zhou2022generalized}            & 87.2          & 88.7          & 89.2          & 88.8          & 89.5          & 90.7          & 76.7          & 80.9          & \textbf{84.7} & 88.4          & 91.5          & 92.1          \\
MGCA–ViT  \cite{NEURIPS2022_d925bda4}             & \textbf{88.8} & \underline{89.1}    & \underline{89.7}    & \underline{89.1}    & \textbf{89.9} & \underline{90.8}    & \textbf{78.9} & \textbf{82.1} & 83.5          & \underline{90.3}    & \underline{92.0}      & \underline{92.6}    \\
MRM \cite{zhou2023advancing}              & 82.7          & 86.3          & 87.0            & 85.2          & 89.2          & 90.4          & 78.3          & 81.7          & 82.4          & 86.7          & 91.3          & 92.3          \\
M3AE \cite{chen2024mapping}        & 84.0            & 86.4          & 88.9          & 86.7          & 88.0            & 89.5          & 74.2          & 79.0            & 80.8          & 86.1          & 88.7          & 90.4          \\
Ours                     & 88.0            & \textbf{89.3} & 89.6          & \textbf{89.3} & \underline{89.6}    & \textbf{90.9} & \underline{78.5}    & \underline{81.8}    & \underline{83.6}    & \textbf{90.9} & \textbf{92.1} & \textbf{92.9}\\
\bottomrule
\end{tabular}
\begin{tablenotes}
    \footnotesize
    \item {AUROC (\%) is used as the classification metric. The \textbf{best} results are bolded and the \underline{second-best} results are underlined.}
\end{tablenotes}
\end{threeparttable}
\end{table*}

\subsection{Cross-modal Retrieval}
The results of cross-modal retrieval are shown in Table IV. Our method achieved the overall best results and achieved the best results in  6 out of 7 evaluation metrics. In addition, our method outperformed the state-of-the-art method MGCA \cite{NEURIPS2022_d925bda4} by 6.5\% at P@1, 11.2\% at P@5, 13.1\% at P@10 on text-to-image retrieval, and 33.9\% at P@Sum. 

\begin{table}[]
\centering
\caption{Cross-modal Retrieval Results on the MIMIC-5x200 Dataset}
\label{table4}
\setlength{\tabcolsep}{3pt}
\begin{threeparttable}
\begin{tabular}{lccc|ccc|c}
\toprule
\multirow{2}{*}{Methods}            & \multicolumn{3}{c|}{Image-to-Text}             & \multicolumn{3}{c|}{Text-to-Image}             & \multirow{2}{*}{P@Sum} \\
                                    & P@1           & P@5           & P@10          & P@1           & P@5           & P@10          &                        \\
                                    \midrule
Random Init                         & 20.0          & 20.0          & 19.8          & 18.0          & 17.3          & 17.7          & 112.8                  \\
ImageNet Init                       & 20.1          & 19.4          & 19.9          & 22.8          & 20.2          & 19.0          & 121.4                  \\
\midrule
\textit{\textbf{CNN-based}}         &               &               &               &               &               &               &                        \\
ConVIRT \cite{zhang2022contrastive}                     & 65.0          & 59.7          & 57.0          & 71.3          & 64.0          & 60.8          & 377.8                  \\
GLoRIA \cite{huang2021gloria}                      & \underline{69.5}          & 60.2          & 57.3          & 73.1          & 63.1          & 59.1          & 382.3                  \\
SAT \cite{liu2023improving}                        & 68.5          & 59.8          & 56.3          & 73.5          & 65.1          & 61.8          & 385.0                  \\
\midrule
\textit{\textbf{Transformer-based}} &               &               &               &               &               &               &                        \\
GLoRIA-ViT  \cite{huang2021gloria}                        & \textbf{69.7} & 61.9          & 58.5          & \underline{74.6}         & 65.3          & 62.1          & 392.1                  \\
MGCA-ViT \cite{NEURIPS2022_d925bda4}                    & 69.1          & \underline{62.5}          & \underline{59.6}          & 73.3          & \underline{66.0}          & \underline{62.7}         & \underline{393.2}                  \\
Ours                                & 67.1          & \textbf{64.0} & \textbf{63.2} & \textbf{79.8} & \textbf{77.2} & \textbf{75.8} & \textbf{427.1}    \\
\bottomrule
\end{tabular}
\begin{tablenotes}
    \footnotesize
    \item {The \textbf{best} results are bolded and the \underline{second-best} results are underlined.}
\end{tablenotes}
\end{threeparttable}
\end{table}

\subsection{Ablation Study}
The first ablation study was conducted on the SIIM dataset for evaluating segmentation and on the MIMIC-5x200 dataset for evaluating the zero-shot classification. This is because our method attempted to explore more fine-grained and accurate text guidance. This guidance can be better evaluated with segmentation and image-text alignment tasks, such that allows to better evaluate the effectiveness of the proposed method. The ablation study results are presented in Table V. Both SRM and IRM boosted the overall performance, where SRM improved more on zero-shot classification results while IRM improved more on segmentation results. Our RECLF combined both SRM and IRM achieved the best results on both tasks.

The second ablation study was performed using the MIMIC-5x200 dataset with zero-shot classification to assess the effectiveness of the multi-block similarity calculation. The results are presented in Table AI in the Appendix. We observed improved results when using vectorized similarity calculations, compared to non-vectorized calculations with $k=1$. Experimentally, we determined that the best results were obtained when $k=12$.

The third ablation study was conducted on the SIIM dataset (segmentation) and the MIMIC-5x200 dataset (zero-shot classification) to assess the impact of various global and local loss configurations. The results are presented in Table AIII in the Appendix. The model had the lowest performance in both segmentation and zero-shot classification tasks when trained with only the global loss. In contrast, RECLF trained with the local loss showed improved performance, particularly in the segmentation task. Ultimately, our RECLF which utilizes both local and global losses achieved the best results.

\begin{table}[]
\centering
\caption{Ablation Study Results}
\label{table5}
\setlength{\tabcolsep}{9pt}
\begin{threeparttable}
\begin{tabular}{ccccc|cc}
\toprule
\multirow{2}{*}{SRM} & \multirow{2}{*}{IRM} & \multicolumn{3}{c|}{Segmentation}              & \multicolumn{2}{c}{Z-S Classification} \\
                     &                     & 1\%           & 10\%          & 100\%         & AUROC              & F1                \\
                     \midrule
        \xmark         & \xmark                   & 59.2          & 63.0          & 66.4          & 0.84               & 0.63              \\
                               \cmark  & \xmark                    & 62.8          & 65.1          & 68.0          & 0.87               & 0.66              \\
                     \xmark         & \cmark                      & \textbf{67.1} & 65.8          & 68.7          & 0.85               & 0.65              \\
                     \cmark         & \cmark                    & 66.9          & \textbf{68.5} & \textbf{69.3} & \textbf{0.88}      & \textbf{0.67}    \\
                     \bottomrule
\end{tabular}
\begin{tablenotes}
    \footnotesize
    \item {Segmentation results are evaluated with Dice similarity coefficient. Z-S represents Zero-shot. The \textbf{best} results are highlighted using bold.}
\end{tablenotes}
\end{threeparttable}
\end{table}

\subsection{Qualitative Results and Analysis}
We present the t-SNE plots for all the samples in the MIMIC-5x200 dataset, which is shown in Fig. 3. The image representations encoded by the image encoder initialized with ImageNet pretrained weights are not separable across different classes (Fig. 3b). This is likely caused by the large semantic gap between natural images and medical images and demonstrates the importance of learning image representations tailored for medical images. All other mVLCL methods achieved competitive performance (Fig. 3c-e). However, as shown in Fig. 3c, GLoRIA-ViT fails to separate images of Atelectasis and Cardiomegaly. In contrast, MGCA-ViT (Fig. 3d) and our RECLF (Fig. 3e) manage to cluster all classes, and our method is slightly better at clustering Cardiomegaly images. We attribute this to the capability in better encoding the critical semantic information in the image representations.

\begin{figure*}[!t]
\centerline{\includegraphics[width=\textwidth]{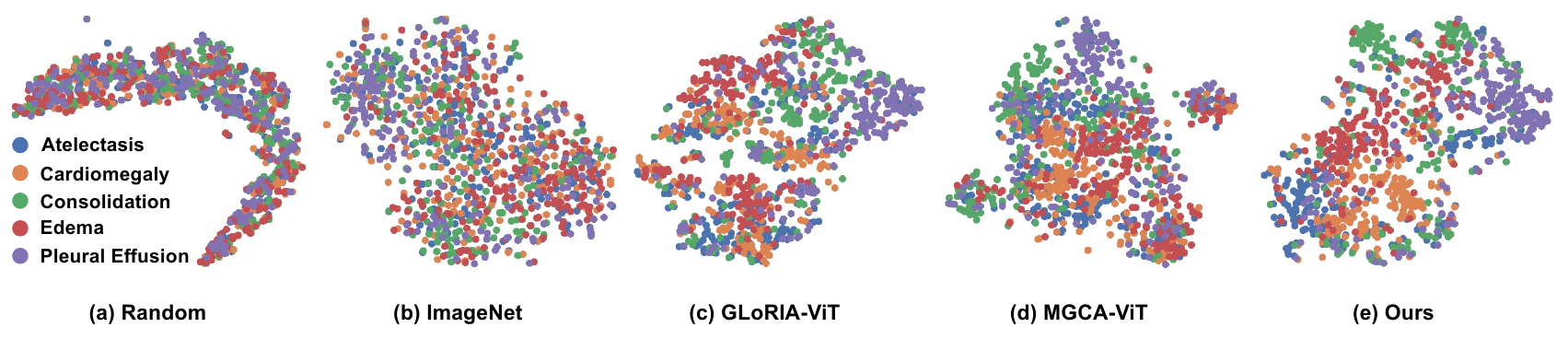}}
\caption{t-SNE visualization of encoded global image representations on the MIMIC-5x200 dataset and the image encoders were initialized with different weights, including: (a) random weights, (b) ImageNet pretrained weights, (c) GLoRIA pretrained weights, (d) MGCA-ViT pretrained weights, and (e) our RECLF pretrained weights. Different colors indicate different class labels.}
\label{fig3}
\end{figure*}

We qualitatively evaluate the learned token-correspondence of our RECLF through attention maps as shown in Fig. 4. For a given word, corresponding image sub-regions were highlighted. Here, our method correctly localized critical image regions for key medical terms such as “effusion” where the heatmap concentrated on the lower part of the chest and the concentration is associated with the collection of fluid.

\begin{figure*}[!t]
\centerline{\includegraphics[width=\textwidth]{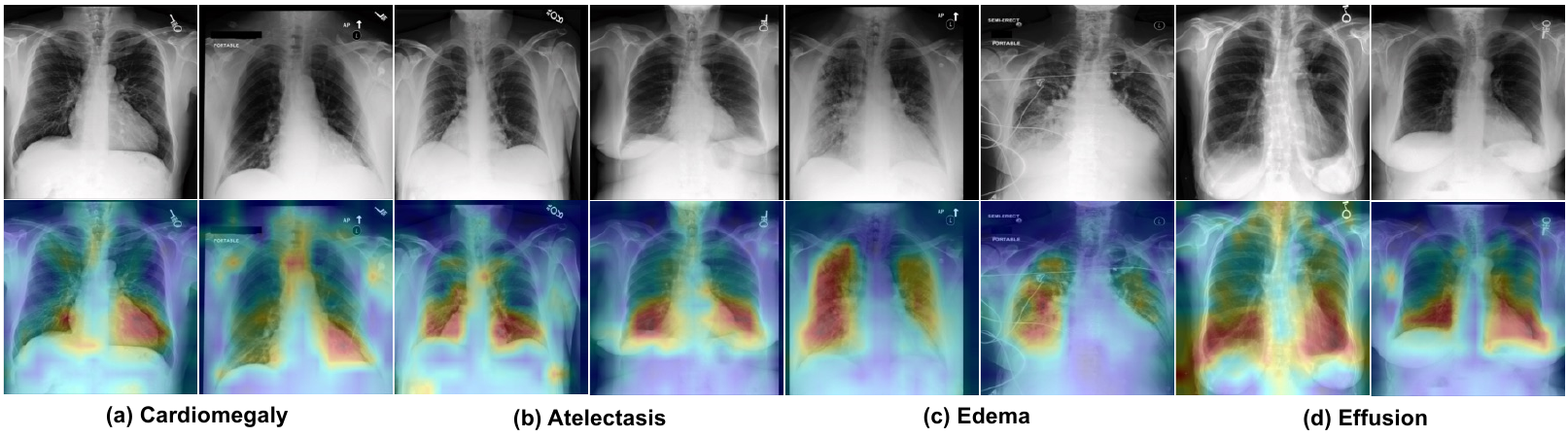}}
\caption{Visualization of learned attention map for a given word. The input images are shown in the 1st row and the learned attention map are shown in the 2nd row. The attention values are normalized to 0-1, which was then mapped to the ‘jet’ lookup table. Red color represents the highest and blue color represents the lowest.}
\label{fig4}
\end{figure*}

Fig. 5 shows example results of image-to-text and text-to-image retrieval of our method and MGCA. MGCA was selected as it was the second-best performing method. The results showed that our method can effectively retrieved the most relevant candidates from the repository when compared to MGCA. Moreover, the retrieved candidates with blue labels indicate that there exist strong semantic similarities between images and text reports from different image-text pairs (for different patients) if they are describing the same disease type.

\begin{figure*}[!t]
\centerline{\includegraphics[width=\textwidth]{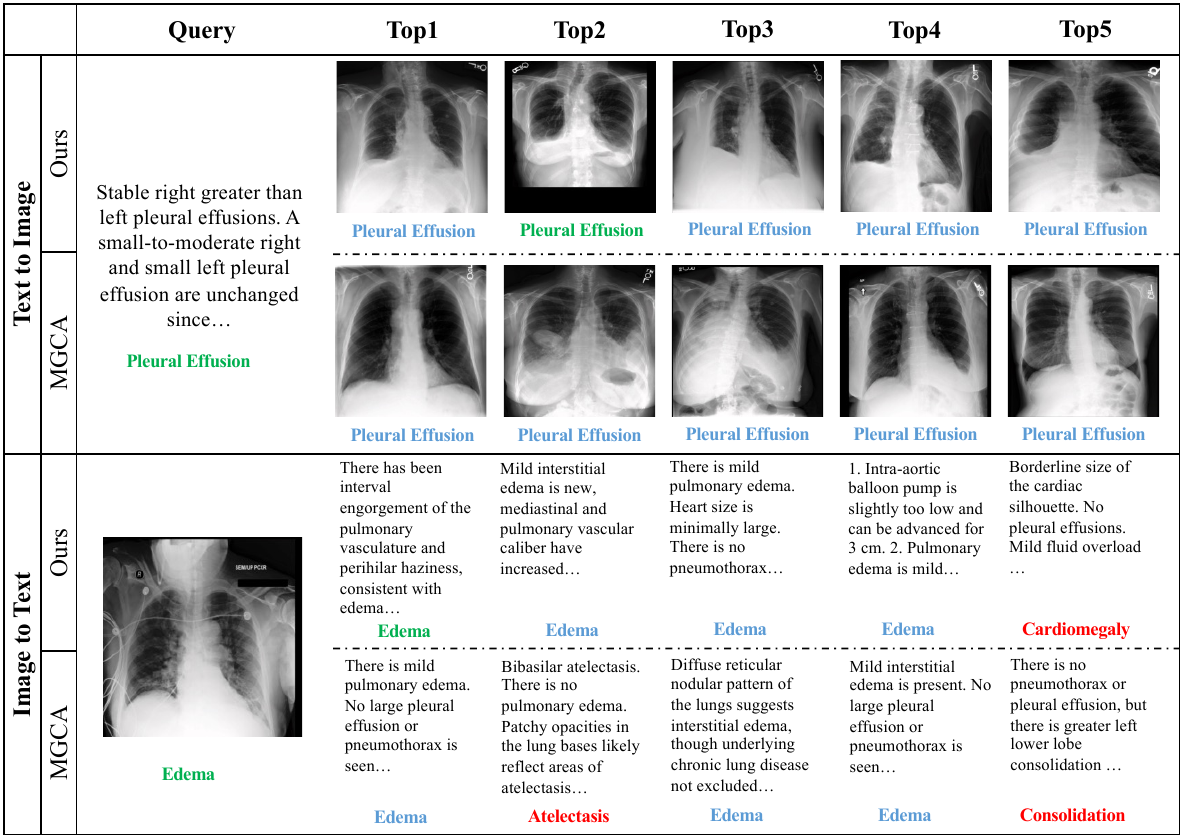}}
\caption{An example text-to-image and image-to-text retrieval results of our method and the comparison MGCA method on the MIMIC-5x200 dataset. For each query, we present the top 5 retrieved results (from left to right). The associated labels are colored with green for a correct match (query and results are from the same patient), or with blue for semantic similar match (the results are describing the same disease but derived from different patients) or with red otherwise (the results are irrelevant). Only important sentences from the text report are reproduced here.}
\label{fig5}
\end{figure*}

In addition, we visualize the learned semantic relations of our SRM through a graph in Fig. A1 in the Appendix. Here, the local-matchings of "left" and "lung" exhibit high semantic relations (Fig. A1f and g). And the local-matching related to "atelectasis" is particularly important, contributing to nearly all other local-matchings (Fig. A1a-g). Notably, the local-matchings related to "minimal," "atelectasis," and "at" demonstrate high semantic relations with those associated with "left" and "lung" (Fig. A1a, c and, d), which illustrates the important disease location information. This demonstrates that our SRM promotes aggregating semantically related local matchings in a unified way by facilitating information propagation among them, and thereby improving the accuracy of similarity measurements.

We also present the t-SNE plots for all samples in the MIMIC-5x200 dataset and the attention map visualizations using different modules of our proposed RECLF, as shown in Fig. A2 and A3 in the Appendix. As expected, employing either SRM or IRM leads to more separated image representations in the latent space compared to the baselines without SRM or IRM (Fig. A2). For example, representations of Cardiomegaly are more clustered when using SRM or IRM (Fig. A2c) compared to baseline (Fig. A2a). Furthermore, the Atelectasis is more clustered when employing SRM and IRM simultaneously (Fig. A2d). We attribute this to the enhanced capability of our RECELF in encoding the critical semantic information in the image representations through more fine-grained contrastive learning facilitated by SRM and IRM. It is also observed that the learned word attention maps from all the ablation results are able to localize the diseased regions (Fig. A3). These results demonstrate that mVLCL can associate text keywords with critical disease-related regions in images and can allow textual guidance to be used as a form of weak supervision for learning general image representations.

We present the visualization of learned similarity metrics before and after SRM (Fig. A4), and the visualization of learned IRM attention weights across various local matchings (Fig. A5) in the Appendix. As shown in Fig. A1 in the Appendix, local-matchings linked to words like "left," "lung", and "atelectasis" are semantically interconnected. Fig. A4 demonstrates that when matching images with positive samples that accurately describe disease location, SRM enhances the local matching related to "atelectasis," increasing its similarity metrics. Conversely, when faced with negative samples that describe the same disease at opposite locations, SRM integrates information from location-related local matchings, such that reduces the similarity metrics related to "atelectasis." As a result, the final similarity of positive samples is higher than that of hard-negative samples, promoting more accurate contrastive learning and efficient image representation with text guidance. Besides, as shown in Fig. A5, local-matchings related to "residual" and "the" are of lower importance. Our IRM adeptly identifies and emphasizes the crucial local matchings, putting more attention on this vital information. These observations are consistent with our assumptions that semantic and important relations among different local-matchings are useful for more accurate and efficient mVLCL, as in Fig. 1.

\section{Discussion}
Our main findings are: (i) Our RECLF outperformed the comparison mVLCL methods on various downstream tasks by explicitly modelling the semantic and importance inter-matching relations, (ii) our RECLF yielded larger improvements in the segmentation task by employing the text guidance on localized image sub-regions and, (iii) our RECLF outperformed the comparison methods on cross-modal image-text alignment tasks (cross-modal retrieval and zero-shot classification) by promoting more accurate image-text alignment.

Our RECLF outperformed the state-of-the-art methods on segmentation task (Table I), zero-shot classification task (Table II), cross-modal retrieval task (Table IV), and ranked first or second-best on the linear classification task (Table III). We attribute these improvements to inter-matching relation modelling via the proposed SRM and IRM. For the ablation study (Table V), our method outperformed the counterparts that were only trained with semantic loss. Our SRM fused the information across all semantic-related local-matchings and thus promoted more accurate image-text alignment, such that benefited cross-modal related tasks e.g., zero-shot classification. Our IRM further enabled the model to focus on critical local-matchings and thus reduces the distractions from irrelevant image sub-regions, which provides advantages in segmentation task.

Our RECLF yielded larger improvements in the segmentation task (Table I). MRM achieved the second-best result, and this likely attributed to the use an additional image super-resolution module such that constrains model to learn more detailed image local features. Although MGCA achieved great performance on classification related tasks, however, MGCA cannot retain this performance on segmentation. This discrepancy arises because MGCA emphasizes inter-class differences (disease-level) between the images and texts, which makes it particularly suitable for disease-level classification tasks rather than segmentation tasks, which is more localized task. The improvements of our RECLF over comparison methods on segmentation task demonstrate the capabilities in learning localized image features. We attribute this improvement to our SRM promotes text guidance on more accurate image sub-regions by reducing the ambiguity arising from hard negative samples. Further, using IRM to minimize the influence from the background local image patches, our model can be guided towards the image representations of more crucial image patches. This aligns with the aim of image segmentation task, which is to identify the localized pathological regions. 

Our RECLF also outperformed the state-of-the-art methods on image-text alignment-related tasks (Table II, Table IV and Table AII). In these tasks, our SRM promoted more accurate similarity measurements by integrating semantic-related local-matchings while IRM filtered out the less relevant local-matchings. These two modules allow to reinforce the consistency between the image and text semantics, such that their similarities in the common latent space can be optimized. This design aligns with the aim of cross-modal retrieval task that retrieves candidates with the highest similarities, the zero-shot classification task that classifies the image by identifying the class prompt with the highest similarity to the image, and the zero-shot phrase grounding task that locates the disease region with the highest similarity to the text.

Table III shows that our RECLF had the highest results in 6 evaluation metrics on the linear classification task and MGCA ranked the best overall. This is likely because MGCA used additional prototype-prototype (disease-disease) matching to differentiate different classes. Therefore, MGCA places greater emphasis on image-image intra-modal differences, which is beneficial for the linear classification task. In contrast, our RECLF puts more weights on inter-modal image-text distances. Furthermore, our model learns to differentiate different classes and to differentiate hard negative samples within the same class (e.g., the right basal effusion case and the left basal effusion case) that exhibits intra-class differences. Consequently, when evaluated in downstream tasks that emphasize intra-modal inter-class differences, such as linear classification, our RECLF produced a slightly worse result when compared to MGCA. On the contrary, in segmentation where intra-class difference is more important (right basal effusion and left basal effusion are located in different places), our method outperformed MGCA in a large margin. Most mVLCL methods outperformed the image-only self-supervised method – MoCo V2 on linear classification, except for ConVIRT and MedKLIP. The compromised results of MedKLIP are likely due to the lack of pretrained weights for the last network block. This also proves the effectiveness of incorporating weak supervision from text reports for image representation learning. 

Our RECLF, which employs multi-block vectorized similarity calculation for distance modelling within each local and global matching, outperformed the non-vectorized version (Table AI). This can be attributed to the fact that different filters capture distinct characteristics of the data, such as edges, size, and textures, and multi-block similarity calculations enable the model to focus on various feature aspects concurrently rather than compressing all similarity information in a single scalar similarity value. This also aligns with the multi-head attention mechanism in Transformers \cite{vaswani2017attention}, which effectively captures diverse input data relations through different attention heads along the channel dimension. Such fine-grained distance modelling enhances the efficiency of contrastive learning, especially in complex scenarios where the relationships between medical images and textual reports are intricate and nuanced.

Table AIII demonstrates that our RECLF achieves the highest performance in both segmentation and zero-shot classification when utilizing both global and local losses. The performance is more profound in the segmentation task, likely due to the local image-text matching (linked to local loss) which allows our model to apply fine-grained textual guidance onto localized image sub-regions, as opposed to relying solely on global image-text matching (linked to global loss). This is consistent with the findings in GLoRIA \cite{huang2021gloria} where they achieved better results with both global and local losses when compared to using global loss only. Additionally, our proposed SRM and IRM, which depend on local image-text matchings, further improve the outcomes when the local loss is applied. This further enhances the model's results by exploring semantic- and importance- relations among local matchings for more accurate textual guidance.

Across all four downstream tasks, the methods employing local-matchings, including GLoRIA, MGCA, and our RECLF, generally outperformed ConVIRT which only using global-matching. These results illustrate the importance of focusing on crucial image sub-regions when contrasting with keywords in the reports. Moreover, the methods employing Transformer as the image encoder, including our RECLF, showed performance improvements over the ResNet-50 counterparts on most of evaluation metrics. These improvements are also consistent with previous investigation \cite{shamshad2023transformers} where Transformers provide benefit in capturing long-range dependence. Another finding is that our RECLF, MedCLIP, MGCA, and SAT that incorporated soft semantic targets for loss computation resulted in higher performance than GLoRIA and ConVIRT which considered all non-paired samples as negative cases. This can be attributed to that medical images and reports from different image-text pairs (for different patients) could have large similarities, i.e., describing the same disease, thus resulting in potential confusion for effective representation learning. Nevertheless, such confusion can be minimized by refraining from simply categorizing all unpaired image-text as negative samples. 

\section{Conclusion}
We have outlined a relation-enhanced framework for mVLCL. Unlike existing methods which neglect the relations between all local-matchings, our network can explicitly model these relations and thus lead to fine-grained utilization of text guidance. The experimental results show that our inter-matching relation-enhanced framework consistently outperform the state-of-the-art methods across multiple downstream tasks. In our future work, we would like to evaluate our relation-enhanced framework on additional downstream tasks, such as detection and visual question answering. In addition, other relation modelling techniques such as semantic scene graph extraction \cite{jain2021radgraph,yu2020ernievil}, can be incorporated to further boost the performance.

\appendices

\section*{Appendix}

Phrase grounding, an essential task for medical image analysis, aims to locate the most relevant region in a medical image based on a given phrase query that describes specific medical findings. We assessed the zero-shot phrase grounding task using the RSNA Pneumonia dataset. The RSNA Pneumonia dataset contains the bounding box labels of the pneumonia areas and we directly evaluated on the test split without fine-tuning. We computed the attention heatmap by contrasting the global representation of the textual class name "Pneumonia" with all local representations of image sub-regions. The region showing the highest response value was identified and compared with the ground-truth segmentation mask. A prediction is considered positive if this region overlaps with the ground-truth mask; otherwise, it is considered negative. We used the Pointing Game score \cite{zhang2018topneural} as the evaluation metric, with results presented in Table AII in the Appendix. Our method surpasses all other methods in the zero-shot phrase grounding task, demonstrating its robust capability to align textual descriptions with image regions of interest (ROIs) in a fine-grained manner.

In our implementation, we utilize a single-layer graph model for the SRM, which results in an additional 468 model parameters across the three projection layers and approximately 0.5 million floating-point operations (FLOPs) in computation cost. Notably, for the IRM, there are no additional model parameters that require training.

\begin{table}[]
\renewcommand{\thetable}{AI}
\centering
\caption{Ablation Study Results of Hyperparameter $k$}
\label{tableAI}
\setlength{\tabcolsep}{12pt}
\begin{threeparttable}
\begin{tabular}{ccc}
\toprule
\multirow{2}{*}{Dim $k$ of the similarity vector} & \multicolumn{2}{c}{MIMIC-5x200 } \\
                                                  & AUROC                  & F1                    \\
\midrule
1                                                 & 0.84                   & 0.63                  \\
12                                                & 0.88                   & 0.67                  \\
48                                                & 0.87                   & 0.67                  \\
\bottomrule
\end{tabular}
\end{threeparttable}
\end{table}

\begin{table}[]
\renewcommand{\thetable}{AII}
\centering
\caption{Zero-shot Phrase Grounding Results on RSNA Dataset}
\label{tableAII}
\setlength{\tabcolsep}{4pt}
\begin{threeparttable}
\begin{tabular}{lcccc}
\toprule
Methods               & GLoRIA \cite{huang2021gloria} & BioViL \cite{boecking2022making} & MedKLIP \cite{wu2023medklip} & Ours \\
\midrule
Pointing Game & 0.76   & 0.83   & 0.87    & 0.91 \\
\bottomrule
\end{tabular}
\end{threeparttable}
\end{table}

\begin{table}[!]
\renewcommand{\thetable}{AIII}
\centering
\caption{Ablation Study Results of Local and Global Loss of RECLF}
\label{tableAIII}
\setlength{\tabcolsep}{6pt}
\begin{threeparttable}
\begin{tabular}{ccccc|cc}
\toprule
\multirow{2}{*}{$\text{Loss}_{global}$} & \multirow{2}{*}{$\text{Loss}_{local}$} & \multicolumn{3}{c|}{Segmentation} & \multicolumn{2}{c}{Z-S Classification} \\
                        &                        & 1\%       & 10\%     & 100\%     & AUROC              & F1                \\
\midrule
\cmark                       & \xmark                      & 58.7      & 62.4     & 66.5      & 0.87               & 0.63              \\
\xmark                       & \cmark                      & 66.3      & 67.7     & 68.6      & 0.87               & 0.65              \\
\cmark                       & \cmark                      & 66.9      & 68.5     & 69.3      & 0.88               & 0.67             \\
\bottomrule
\end{tabular}
\end{threeparttable}
\end{table}

\begin{figure*}[!t]
\renewcommand{\thefigure}{A1}
\centerline{\includegraphics[width=\textwidth]{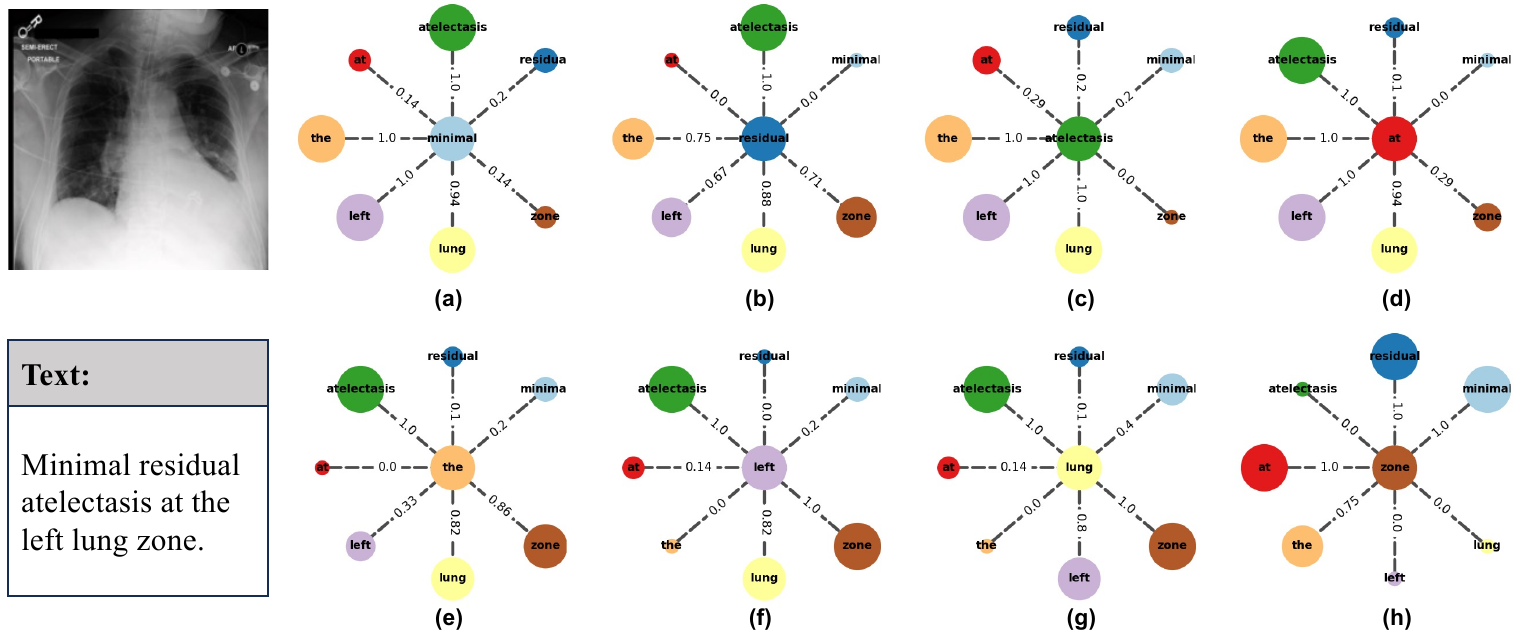}}
\caption{Visualization of the semantic relationships among the local matchings in the SRM's constructed graph. The text is “Minimal residual atelectasis at the left lung zone.” Each subgraph depicts the connections between the selected local matching (central node) and all other local matchings (peripheral nodes). Node labels correspond to the associated text word for each local matching. The edges reflect these semantic relationships, with their weights normalized to a scale of 0 to 1. Higher edge weights suggest stronger semantic connections between the local matchings (nodes). For improved visualization, the size of the peripheral nodes is scaled according to the edge weight.}
\label{figA1}
\end{figure*}

\begin{figure*}[!t]
\renewcommand{\thefigure}{A2}
\centerline{\includegraphics[width=\textwidth]{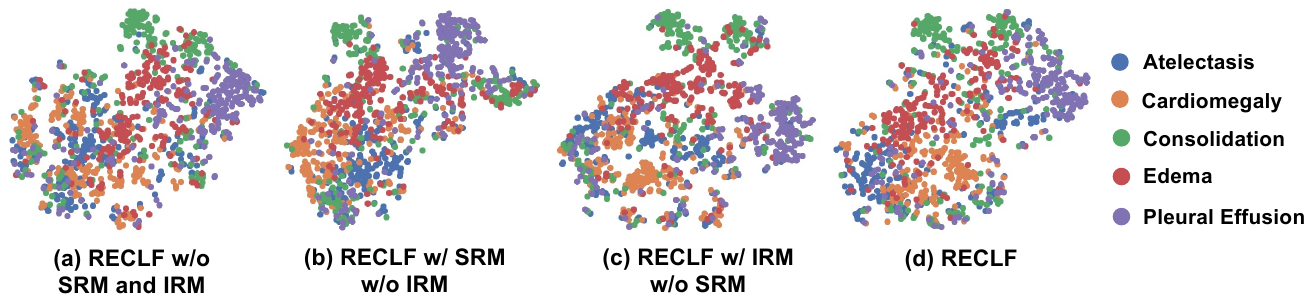}}
\caption{t-SNE visualization of the encoded global image representations on the MIMIC-5x200 dataset and the image encoders which were initialized with different weights: (a) RECLF without SRM and IRM, (b) RECLF with SRM without IRM, (c) RECLF with IRM without SRM, and (d) our RECLF. Different colors indicate different class labels.}
\label{figA2}
\end{figure*}

\begin{figure*}[!t]
\renewcommand{\thefigure}{A3}
\centerline{\includegraphics[width=\textwidth]{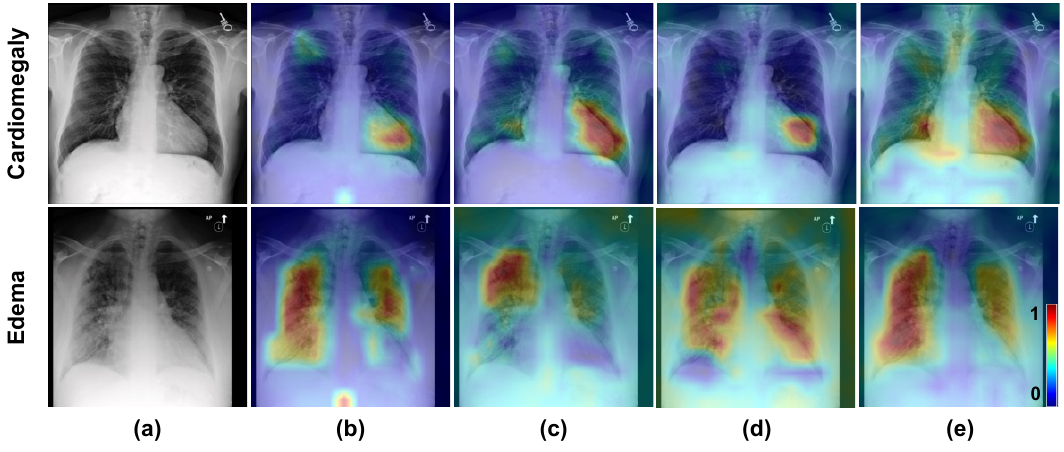}}
\caption{Visualization of learned attention map for a given word using different components of our proposed model: (a) raw image, (b) RECLF without SRM and IRM, (c) RECLF with SRM without IRM, (d) RECLF with IRM without SRM, and (d) our RECLF. The attention values are normalized to 0-1, which was then mapped to the ‘jet’ color lookup table.}
\label{figA3}
\end{figure*}

\begin{figure*}[!t]
\renewcommand{\thefigure}{A4}
\centerline{\includegraphics[width=\textwidth]{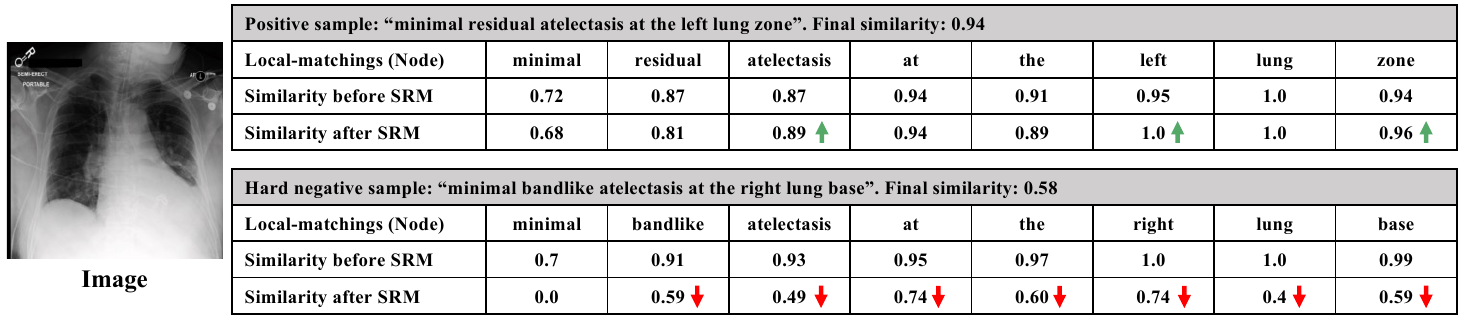}}
\caption{Visualization of learned similarity metrics of different local-matchings before and after SRM using a positive sample and a hard negative sample, respectively.}
\label{figA4}
\end{figure*}

\begin{figure*}[!t]
\renewcommand{\thefigure}{A5}
\centerline{\includegraphics[width=\textwidth]{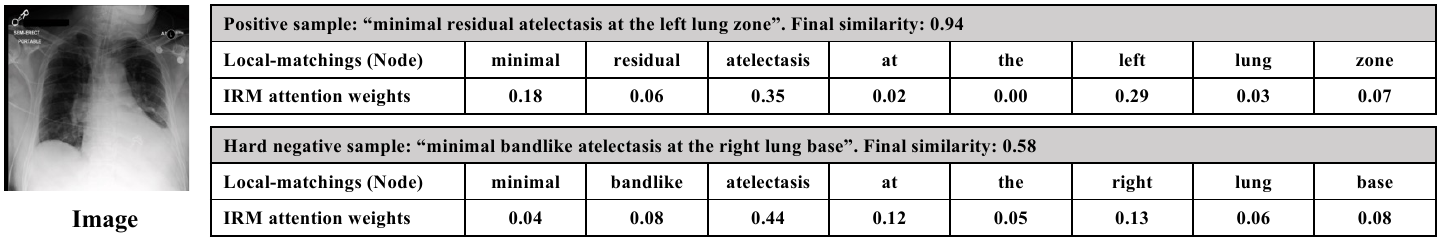}}
\caption{Visualization of learned IRM attention weights of different local-matchings using a positive sample and a hard negative sample, respectively.}
\label{figA5}
\end{figure*}

\bibliographystyle{ieeetr}
\bibliography{tmi}

\begin{thebibliography}{10}

\bibitem{zhang2022contrastive}
Y.~Zhang, H.~Jiang, Y.~Miura, C.~D. Manning, and C.~P. Langlotz, ``Contrastive learning of medical visual representations from paired images and text,'' in {\em Machine {{Learning}} for {{Healthcare Conference}}}, pp.~2--25, {PMLR}, 2022.

\bibitem{tiu2022expertlevel}
E.~Tiu, E.~Talius, P.~Patel, C.~P. Langlotz, A.~Y. Ng, and P.~Rajpurkar, ``Expert-level detection of pathologies from unannotated chest {{X-ray}} images via self-supervised learning,'' {\em Nat. Biomed. Eng}, vol.~6, pp.~1399--1406, Dec. 2022.

\bibitem{moor2023foundation}
M.~Moor, O.~Banerjee, Z.~S.~H. Abad, H.~M. Krumholz, J.~Leskovec, E.~J. Topol, and P.~Rajpurkar, ``Foundation models for generalist medical artificial intelligence,'' {\em Nature}, vol.~616, pp.~259--265, Apr. 2023.

\bibitem{huang2023selfsupervised}
S.-C. Huang, A.~Pareek, M.~Jensen, M.~P. Lungren, S.~Yeung, and A.~S. Chaudhari, ``Self-supervised learning for medical image classification: A systematic review and implementation guidelines,'' {\em npj Digit. Med.}, vol.~6, pp.~1--16, Apr. 2023.

\bibitem{krishnan2022selfsupervised}
R.~Krishnan, P.~Rajpurkar, and E.~J. Topol, ``Self-supervised learning in medicine and healthcare,'' {\em Nat. Biomed. Eng}, vol.~6, pp.~1346--1352, Dec. 2022.

\bibitem{sun2023scoping}
Z.~Sun, M.~Lin, Q.~Zhu, Q.~Xie, F.~Wang, Z.~Lu, and Y.~Peng, ``A scoping review on multimodal deep learning in biomedical images and texts,'' {\em Journal of Biomedical Informatics}, vol.~146, p.~104482, Oct. 2023.

\bibitem{NEURIPS2022_d925bda4}
F.~Wang, Y.~Zhou, S.~WANG, V.~Vardhanabhuti, and L.~Yu, ``Multi-granularity cross-modal alignment for generalized medical visual representation learning,'' in {\em Advances in Neural Information Processing Systems} (S.~Koyejo, S.~Mohamed, A.~Agarwal, D.~Belgrave, K.~Cho, and A.~Oh, eds.), vol.~35, pp.~33536--33549, {Curran Associates, Inc.}, 2022.

\bibitem{huang2021gloria}
S.-C. Huang, L.~Shen, M.~P. Lungren, and S.~Yeung, ``{{GLoRIA}}: {{A Multimodal Global-Local Representation Learning Framework}} for {{Label-Efficient Medical Image Recognition}},'' in {\em Proceedings of the {{IEEE}}/{{CVF International Conference}} on {{Computer Vision}}}, pp.~3942--3951, 2021.

\bibitem{liu2023improving}
B.~Liu, D.~Lu, D.~Wei, X.~Wu, Y.~Wang, Y.~Zhang, and Y.~Zheng, ``Improving {{Medical Vision-Language Contrastive Pretraining}} with {{Semantics-aware Triage}},'' {\em IEEE Transactions on Medical Imaging}, pp.~1--1, 2023.

\bibitem{wang2022medclip}
Z.~Wang, Z.~Wu, D.~Agarwal, and J.~Sun, ``{{MedCLIP}}: {{Contrastive Learning}} from {{Unpaired Medical Images}} and {{Text}},'' in {\em Proceedings of the 2022 {{Conference}} on {{Empirical Methods}} in {{Natural Language Processing}}}, ({Abu Dhabi, United Arab Emirates}), pp.~3876--3887, {Association for Computational Linguistics}, Dec. 2022.

\bibitem{chen2023knowledge}
X.~Chen, Y.~He, C.~Xue, R.~Ge, S.~Li, and G.~Yang, ``Knowledge {{Boosting}}: {{Rethinking Medical Contrastive Vision-Language Pre-training}},'' in {\em Medical {{Image Computing}} and {{Computer Assisted Intervention}} {\textendash} {{MICCAI}} 2023} (H.~Greenspan, A.~Madabhushi, P.~Mousavi, S.~Salcudean, J.~Duncan, T.~{Syeda-Mahmood}, and R.~Taylor, eds.), Lecture {{Notes}} in {{Computer Science}}, ({Cham}), pp.~405--415, {Springer Nature Switzerland}, 2023.

\bibitem{karpathy2015deep}
A.~Karpathy and L.~{Fei-Fei}, ``Deep visual-semantic alignments for generating image descriptions,'' in {\em Proceedings of the {{IEEE}} Conference on Computer Vision and Pattern Recognition}, pp.~3128--3137, 2015.

\bibitem{muller2022joint}
P.~M{\"u}ller, G.~Kaissis, C.~Zou, and D.~Rueckert, ``Joint {{Learning}} of~{{Localized Representations}} from~{{Medical Images}} and~{{Reports}},'' in {\em Computer {{Vision}} {\textendash} {{ECCV}} 2022} (S.~Avidan, G.~Brostow, M.~Ciss{\'e}, G.~M. Farinella, and T.~Hassner, eds.), Lecture {{Notes}} in {{Computer Science}}, ({Cham}), pp.~685--701, {Springer Nature Switzerland}, 2022.

\bibitem{he2016deep}
K.~He, X.~Zhang, S.~Ren, and J.~Sun, ``Deep residual learning for image recognition,'' in {\em Proceedings of the {{IEEE}} Conference on Computer Vision and Pattern Recognition}, pp.~770--778, 2016.

\bibitem{dosovitskiy2020image}
A.~Dosovitskiy, L.~Beyer, A.~Kolesnikov, D.~Weissenborn, X.~Zhai, T.~Unterthiner, M.~Dehghani, M.~Minderer, G.~Heigold, and S.~Gelly, ``An image is worth 16x16 words: {{Transformers}} for image recognition at scale,'' {\em arXiv preprint arXiv:2010.11929}, 2020.

\bibitem{shamshad2023transformers}
F.~Shamshad, S.~Khan, S.~W. Zamir, M.~H. Khan, M.~Hayat, F.~S. Khan, and H.~Fu, ``Transformers in medical imaging: {{A}} survey,'' {\em Medical Image Analysis}, vol.~88, p.~102802, Aug. 2023.

\bibitem{chen2022multimodal}
Z.~Chen, Y.~Du, J.~Hu, Y.~Liu, G.~Li, X.~Wan, and T.-H. Chang, ``Multi-modal {{Masked Autoencoders}} for~{{Medical Vision-and-Language Pre-training}},'' in {\em Medical {{Image Computing}} and {{Computer Assisted Intervention}} {\textendash} {{MICCAI}} 2022} (L.~Wang, Q.~Dou, P.~T. Fletcher, S.~Speidel, and S.~Li, eds.), Lecture {{Notes}} in {{Computer Science}}, ({Cham}), pp.~679--689, {Springer Nature Switzerland}, 2022.

\bibitem{ghorbani2022ragcn}
M.~Ghorbani, A.~Kazi, M.~Soleymani~Baghshah, H.~R. Rabiee, and N.~Navab, ``{{RA-GCN}}: {{Graph}} convolutional network for disease prediction problems with imbalanced data,'' {\em Medical Image Analysis}, vol.~75, p.~102272, Jan. 2022.

\bibitem{li2022satr}
H.~Li, L.~Chen, H.~Han, and S.~Kevin~Zhou, ``{{SATr}}: {{Slice Attention}} with {{Transformer}} for {{Universal Lesion Detection}},'' in {\em Medical {{Image Computing}} and {{Computer Assisted Intervention}} {\textendash} {{MICCAI}} 2022} (L.~Wang, Q.~Dou, P.~T. Fletcher, S.~Speidel, and S.~Li, eds.), vol.~13433, pp.~163--174, {Cham}: {Springer Nature Switzerland}, 2022.

\bibitem{li2022dual}
Y.~Li, Y.~Zhang, W.~Cui, B.~Lei, X.~Kuang, and T.~Zhang, ``Dual {{Encoder-Based Dynamic-Channel Graph Convolutional Network With Edge Enhancement}} for {{Retinal Vessel Segmentation}},'' {\em IEEE Transactions on Medical Imaging}, vol.~41, pp.~1975--1989, Aug. 2022.

\bibitem{nakao2022imagetograph}
M.~Nakao, M.~Nakamura, and T.~Matsuda, ``Image-to-{{Graph Convolutional Network}} for {{2D}}/{{3D Deformable Model Registration}} of {{Low-Contrast Organs}},'' {\em IEEE Transactions on Medical Imaging}, vol.~41, pp.~3747--3761, Dec. 2022.

\bibitem{vaswani2017attention}
A.~Vaswani, N.~Shazeer, N.~Parmar, J.~Uszkoreit, L.~Jones, A.~N. Gomez, {\textbackslash}.~Kaiser, and I.~Polosukhin, ``Attention is all you need,'' in {\em Advances in {{Neural Information Processing Systems}}}, pp.~5998--6008, 2017.

\bibitem{devlin2019bert}
J.~Devlin, M.-W. Chang, K.~Lee, and K.~Toutanova, ``{{BERT}}: {{Pre-training}} of {{Deep Bidirectional Transformers}} for {{Language Understanding}},'' May 2019.

\bibitem{alsentzer2019publicly}
E.~Alsentzer, J.~R. Murphy, W.~Boag, W.-H. Weng, D.~Jin, T.~Naumann, and M.~B.~A. McDermott, ``Publicly {{Available Clinical BERT Embeddings}},'' June 2019.

\bibitem{boecking2022making}
B.~Boecking, N.~Usuyama, S.~Bannur, D.~C. Castro, A.~Schwaighofer, S.~Hyland, M.~Wetscherek, T.~Naumann, A.~Nori, and J.~{Alvarez-Valle}, ``Making the {{Most}} of {{Text Semantics}} to {{Improve Biomedical Vision}}{\textendash}{{Language Processing}},'' {\em arXiv preprint arXiv:2204.09817}, 2022.

\bibitem{wu2023medklip}
C.~Wu, X.~Zhang, Y.~Zhang, Y.~Wang, and W.~Xie, ``{{MedKLIP}}: {{Medical Knowledge Enhanced Language-Image Pre-Training}} for {{X-ray Diagnosis}},'' in {\em Proceedings of the {{IEEE}}/{{CVF International Conference}} on {{Computer Vision}} ({{ICCV}})}, pp.~21372--21383, Oct. 2023.

\bibitem{zhang2023knowledgeenhanced}
X.~Zhang, C.~Wu, Y.~Zhang, W.~Xie, and Y.~Wang, ``Knowledge-enhanced visual-language pre-training on chest radiology images,'' {\em Nat Commun}, vol.~14, p.~4542, July 2023.

\bibitem{jain2021radgraph}
S.~Jain, A.~Agrawal, A.~Saporta, S.~Truong, T.~Bui, P.~Chambon, Y.~Zhang, M.~P. Lungren, A.~Y. Ng, and C.~Langlotz, ``{{RadGraph}}: {{Extracting Clinical Entities}} and {{Relations}} from {{Radiology Reports}},'' in {\em Thirty-Fifth {{Conference}} on {{Neural Information Processing Systems Datasets}} and {{Benchmarks Track}} ({{Round}} 1)}, 2021.

\bibitem{zhao2023clip}
Z.~Zhao, Y.~Liu, H.~Wu, Y.~Li, S.~Wang, L.~Teng, D.~Liu, Z.~Cui, Q.~Wang, and D.~Shen, ``{{CLIP}} in {{Medical Imaging}}: {{A Comprehensive Survey}},'' Dec. 2023.

\bibitem{wan2023medunic}
Z.~Wan, C.~Liu, M.~Zhang, J.~Fu, B.~Wang, S.~Cheng, L.~Ma, C.~{Quilodr{\'a}n-Casas}, and R.~Arcucci, ``Med-{{UniC}}: {{Unifying Cross-Lingual Medical Vision-Language Pre-Training}} by {{Diminishing Bias}},'' in {\em Thirty-Seventh {{Conference}} on {{Neural Information Processing Systems}}}, Nov. 2023.

\bibitem{liu2020graph}
C.~Liu, Z.~Mao, T.~Zhang, H.~Xie, B.~Wang, and Y.~Zhang, ``Graph structured network for image-text matching,'' in {\em Proceedings of the {{IEEE}}/{{CVF Conference}} on {{Computer Vision}} and {{Pattern Recognition}}}, pp.~10921--10930, 2020.

\bibitem{diao2021similarity}
H.~Diao, Y.~Zhang, L.~Ma, and H.~Lu, ``Similarity {{Reasoning}} and {{Filtration}} for {{Image-Text Matching}},'' in {\em Proceedings of the {{AAAI Conference}} on {{Artificial Intelligence}}}, vol.~35, pp.~1218--1226, 2021.

\bibitem{velickovic2018graph}
P.~Veli{\v c}kovi{\'c}, G.~Cucurull, A.~Casanova, A.~Romero, P.~Li{\`o}, and Y.~Bengio, ``Graph {{Attention Networks}},'' in {\em International {{Conference}} on {{Learning Representations}}}, 2018.

\bibitem{johnson2019mimiccxr}
A.~E. Johnson, T.~J. Pollard, S.~J. Berkowitz, N.~R. Greenbaum, M.~P. Lungren, C.-y. Deng, R.~G. Mark, and S.~Horng, ``{{MIMIC-CXR}}, a de-identified publicly available database of chest radiographs with free-text reports,'' {\em Scientific data}, vol.~6, no.~1, p.~317, 2019.

\bibitem{siim-acr-pneumothorax-segmentation}
A.~Zawacki, C.~Wu, G.~Shih, J.~Elliott, M.~Fomitchev, M.~Hussain, {ParasLakhani}, P.~Culliton, and S.~Bao, ``{{SIIM-ACR}} pneumothorax segmentation.''

\bibitem{shih2019augmenting}
G.~Shih, C.~C. Wu, S.~S. Halabi, M.~D. Kohli, L.~M. Prevedello, T.~S. Cook, A.~Sharma, J.~K. Amorosa, V.~Arteaga, M.~{Galperin-Aizenberg}, R.~R. Gill, M.~C. Godoy, S.~Hobbs, J.~Jeudy, A.~Laroia, P.~N. Shah, D.~Vummidi, K.~Yaddanapudi, and A.~Stein, ``Augmenting the {{National Institutes}} of {{Health Chest Radiograph Dataset}} with {{Expert Annotations}} of {{Possible Pneumonia}},'' {\em Radiology: Artificial Intelligence}, vol.~1, p.~e180041, Jan. 2019.

\bibitem{liu2020rethinking}
Y.~Liu, Y.-H. Wu, Y.~Ban, H.~Wang, and M.-M. Cheng, ``Rethinking computer-aided tuberculosis diagnosis,'' in {\em Proceedings of the {{IEEE}}/{{CVF}} Conference on Computer Vision and Pattern Recognition}, pp.~2646--2655, 2020.

\bibitem{zheng2021rethinking}
S.~Zheng, J.~Lu, H.~Zhao, X.~Zhu, Z.~Luo, Y.~Wang, Y.~Fu, J.~Feng, T.~Xiang, and P.~H. Torr, ``Rethinking semantic segmentation from a sequence-to-sequence perspective with transformers,'' in {\em Proceedings of the {{IEEE}}/{{CVF}} Conference on Computer Vision and Pattern Recognition}, pp.~6881--6890, 2021.

\bibitem{irvin2019chexpert}
J.~Irvin, P.~Rajpurkar, M.~Ko, Y.~Yu, S.~{Ciurea-Ilcus}, C.~Chute, H.~Marklund, B.~Haghgoo, R.~Ball, and K.~Shpanskaya, ``Chexpert: {{A}} large chest radiograph dataset with uncertainty labels and expert comparison,'' in {\em Proceedings of the {{AAAI}} Conference on Artificial Intelligence}, vol.~33, pp.~590--597, 2019.

\bibitem{wang2017chestxray8}
X.~Wang, Y.~Peng, L.~Lu, Z.~Lu, M.~Bagheri, and R.~M. Summers, ``Chestx-ray8: {{Hospital-scale}} chest x-ray database and benchmarks on weakly-supervised classification and localization of common thorax diseases,'' in {\em Proceedings of the {{IEEE}} Conference on Computer Vision and Pattern Recognition}, pp.~2097--2106, 2017.

\bibitem{he2021masked}
K.~He, X.~Chen, S.~Xie, Y.~Li, P.~Doll{\'a}r, and R.~Girshick, ``Masked {{Autoencoders Are Scalable Vision Learners}},'' {\em arXiv preprint arXiv:2111.06377}, 2021.

\bibitem{zhou2023advancing}
H.-Y. Zhou, C.~Lian, L.~Wang, and Y.~Yu, ``Advancing {{Radiograph Representation Learning}} with {{Masked Record Modeling}},'' in {\em The {{Eleventh International Conference}} on {{Learning Representations}}}, 2023.

\bibitem{chen2024mapping}
Z.~Chen, Y.~Du, J.~Hu, Y.~Liu, G.~Li, X.~Wan, and T.-H. Chang, ``Mapping medical image-text to a joint space via masked modeling,'' {\em Medical Image Analysis}, vol.~91, p.~103018, Jan. 2024.

\bibitem{chen2020improved}
X.~Chen, H.~Fan, R.~Girshick, and K.~He, ``Improved baselines with momentum contrastive learning,'' {\em arXiv preprint arXiv:2003.04297}, 2020.

\bibitem{Chen_2023_ICCV}
Z.~Chen, S.~Diao, B.~Wang, G.~Li, and X.~Wan, ``Towards unifying medical vision-and-language pre-training via soft prompts,'' in {\em Proceedings of the {{IEEE}}/{{CVF}} International Conference on Computer Vision ({{ICCV}})}, pp.~23403--23413, Oct. 2023.

\bibitem{zhou2022generalized}
H.-Y. Zhou, X.~Chen, Y.~Zhang, R.~Luo, L.~Wang, and Y.~Yu, ``Generalized radiograph representation learning via cross-supervision between images and free-text radiology reports,'' {\em Nature Machine Intelligence}, vol.~4, pp.~32--40, Jan. 2022.

\bibitem{yu2020ernievil}
F.~Yu, J.~Tang, W.~Yin, Y.~Sun, H.~Tian, H.~Wu, and H.~Wang, ``Ernie-vil: {{Knowledge}} enhanced vision-language representations through scene graph,'' {\em arXiv preprint arXiv:2006.16934}, 2020.

\bibitem{zhang2018topneural}
J.~Zhang, S.~A. Bargal, Z.~Lin, J.~Brandt, X.~Shen, and S.~Sclaroff, ``Top-down neural attention by excitation backprop,'' {\em International Journal of Computer Vision}, vol.~126, no.~10, pp.~1084--1102, 2018.

\end{thebibliography}

\end{document}